\documentclass[final,12pt]{elsarticle}
\usepackage{amsmath,amsfonts}
\usepackage{algorithmic}
\usepackage{algorithm}
\usepackage{array}
\usepackage[caption=false,font=scriptsize,labelfont=sf,textfont=sf]{subfig}
\usepackage{textcomp}
\usepackage{stfloats}
\usepackage{url}
\usepackage{verbatim}
\usepackage{graphicx}
\usepackage{framed,xcolor}

\usepackage{multirow}
\usepackage[hidelinks]{hyperref}
\usepackage{booktabs}
\usepackage{orcidlink}
\usepackage{arydshln} 

\definecolor{shadecolor}{gray}{0.9}
\journal{preprint}

\begin{document}

\begin{frontmatter}

\title{Vision-Based Risk Aware Emergency Landing for UAVs in Complex Urban Environments}\corref{fund}

 \author[cimat]{Julio de la Torre-Vanegas}
 \ead{julio.delatorre@cimat.mx}
 \author[cimat]{Miguel Soriano-García}
  \ead{miguel.garcia@cimat.mx}
\address[cimat]{{Center for Research in Mathematics CIMAT AC, campus Zacatecas},
           {Calle Lasec y Andador Galileo Galilei, Manzana 3, Lote 7 Quantum Ciudad del Conocimiento}, 
            {Zacatecas},
            {98160}, 
            {Zacatecas},
            {Mexico}
            }    

            \author[isra]{Israel Becerra}
            \ead{israelb@cimat.mx}
     \address[isra]{{Investigadores por México at Center for Research in Mathematics CIMAT AC}, 
            {calle Jalisco s/n, Valenciana}, 
            {Guanajuato},
            {36023}, 
            {Guanajuato},
            {Mexico}
            }     
\author[die,cor]{Diego Mercado-Ravell}
\ead{diego.mercado@cinvestav.mx}
\address[die]{{Center for Research and Advanced Studies CINVESTAV-IPN, campus Guadalajara},
            {Av. del Bosque 1145, El Bajío}, 
            {Zapopan},
            {45017}, 
            {Jalisco},
            {Mexico}}
\address[cor]{Corresponding author: 
{diego.mercado@cimat.mx}
}

\cortext[fund]{This work was supported by the Office of Naval Research Global ONRG, Award No. N62909-24-1-2001.    }
\begin{abstract}
Landing safely in crowded urban environments remains an essential yet challenging endeavor for Unmanned Aerial Vehicles (UAVs), especially in emergency situations. In this work, we propose a risk-aware approach that harnesses semantic segmentation to continuously evaluate potential hazards in the drone's field of view. By using a specialized deep neural network to assign pixel-level risk values and applying an algorithm based on risk maps, our method adaptively identifies a stable Safe Landing Zone (SLZ) despite moving critical obstacles such as vehicles, people, etc., and other visual challenges like shifting illumination. A control system then guides the UAV toward this low-risk region, employing altitude-dependent safety thresholds and temporal landing point stabilization to ensure robust descent trajectories. Experimental validation in diverse urban environments demonstrates the effectiveness of our approach, achieving over $90\%$ landing success rates in very challenging real scenarios, showing significant improvements in various risk metrics. Our findings suggest that risk-oriented vision methods can effectively help reduce the risk of accidents in emergency landing situations, particularly in complex, unstructured, urban scenarios, densely populated with moving risky obstacles, while potentiating the true capabilities of UAVs in complex urban operations.
\end{abstract}


\end{frontmatter}

\section{Introduction}\label{sec:intro}

Unmanned Aerial Vehicles (UAVs) play an increasingly significant role in civilian applications, facilitating tasks ranging from infrastructure inspection and package delivery to precision agriculture and search-and-rescue operations \cite{Shakhatreh2019,Katkuri2024}. Nevertheless, one major challenge persists: ensuring a safe and reliable landing in unstructured urban environments \cite{Shah2021}. When a system failure or sudden emergency arises, such as battery shortage, communication loss, or mechanical or electronic damage, traditional return-to-home strategies or marker-based landing approaches \cite{Pieczynski2024, Demirhan2020, lv2024autonomous} may not be sufficient, particularly if the intended final landing spot becomes unsafe, unreachable, or unavailable \cite{Morando2024, Haoran2023}.

Regulatory frameworks, including the Specific Operations Risk Assessment (SORA) of the European Union Aviation Safety Agency (EASA) \cite{SORA2019}, reflect these concerns by mandating reliable safeguards for urban UAV operations. Thus, recent efforts have highlighted the importance of context awareness in the design of robust vision-based landing strategies, where semantic segmentation networks arise as an interesting approach \cite{Loera2024, Abdollahzadeh2022, Benjwal2023}. By using Deep Neural Networks (DNN) for classifying the environment into different semantic classes (such as roads, sidewalks, buildings, or vegetation), a UAV can infer real-time contextual cues regarding the potential risk in each region of the scene. For instance, landing on a crowded sidewalk or near fast-moving vehicles poses a higher threat to both people and the drone, whereas an open park or an empty segment of road might be more appropriate. Moreover, the use of context-aware strategies allows for more sophisticated solutions in the decision-making process for autonomous landing, which may be crucial in emergency landing situations during operation in complex urban environments, where we refer to ``complex'' as unstructured urban scenarios, densely populated in presence of high-risk moving obstacles such as people or vehicles. In this context, risk-aware emergency landing poses great potential to enhance the applicability of UAVs in urban operations.

In this work, we provide a complete, fully autonomous landing solution in complex urban scenarios, based only on visual information from a downward-facing monocular camera. First, a semantic segmentation network provides pixel-wise contextual information, which is in turn converted to a risk map. This risk map is directly stored to create a global risk map using the images seen from the start of the landing mission, accumulating the highest perceived risk over time in a kind of persistent “memory” of the detected dangers. Then, image processing algorithms are applied to the local drone's view of this global risk map to obtain a stable risk chart, from which a minimization problem is solved considering the lowest risk at the shortest distance. Subsequently, the UAV is directed toward the selected landing spot as long as it remains available; otherwise, the next best spot is selected. The proposed strategy was thoroughly evaluated in several challenging real urban scenarios employing the VIVA-SAFELAND validation freeware \cite{Soriano-Arxiv2025} (a tool that employs pre-recorded real-life videos to allow an Emulated Aerial Vehicle to navigate using visual cues), where specific quantitative metrics were obtained to assess the performance of the autonomous landing solution.

The outline of the paper is as follows. Sec. \ref{sec:related} presents the literature review for autonomous landing in unstructured environments. Then, Secs. \ref{sec:prob} and~\ref{sec:risk} present the problem definition and the proposed method, respectively. Then, in Sec. \ref{sec:experiment}, the algorithms are evaluated under different real urban scenarios, showcasing their good performance. Finally, conclusions and future work are discussed in Sec. \ref{sec:conclusion}.
%


    \section{Related Work}
    \label{sec:related}
    
    Ensuring safe autonomous landing for Unmanned Aerial Vehicles (UAVs) remains a critical challenge, particularly during system failures or in GPS-denied urban environments where accidents can have severe consequences \cite{Shah2021}. Vision-based systems offer a promising avenue not only due to the rich contextual information cameras readily provide but also because cameras are nearly ubiquitous on UAV platforms, making vision an accessible sensing modality. Previous vision-based strategies for landing in unstructured environments often rely predominantly on geometric features. For instance, \cite{Yang2018} leveraged monocular vision and Simultaneous Localization and Mapping (SLAM) to create 3D maps for identifying potentially clear areas. Similarly, \cite{Kaljahi2019} proposed detecting planarity using Gabor transforms without prior classification, aiming for adaptability. However, these geometric-centric methods often face challenges in the highly dynamic conditions of complex urban environments. Crucially, they typically operate without a contextual understanding of the scene, making it difficult to assess the true nature and inherent risks of an area beyond its immediate geometric properties; their performance can also be sensitive to abrupt UAV movements, poor textures, or varying illumination conditions. Systems relying solely on alternative sensors like LiDAR \cite{Saldiran2024, Saldiran2025} offer robustness to lighting variations and can provide precise geometric data for assessing surface suitability and maintaining continuous verification. However, such systems lack an inherent risk understanding derived from visual cues, which is crucial for interpreting the nature of the terrain and its associated risks (e.g., distinguishing a sidewalk from a road), and depend on sensors that are not universally available on all UAV platforms.
    
    The incorporation of semantic segmentation for Safe Landing Zone (SLZ) identification represents a significant advancement towards richer environmental understanding and smart decision making. While comprehensive reviews like \cite{long2022} have surveyed various vision-based landing techniques, including supervised classifiers and feature extraction, the explicit and widespread integration of semantic segmentation for nuanced risk assessment in landing was, until recently, less common, despite its clear potential. However, ongoing advancements in processing capabilities and hardware are diminishing these barriers, making its application increasingly feasible. Several recent works have indeed begun to incorporate deeper semantic understanding to identify suitable landing areas. For example, \cite{Zhang2025} employed enhanced semantic segmentation with height-dependent classification detail and zone tracking. Other efforts have offered comprehensive frameworks for SLZ selection; for instance, \cite{Wu2024} addressed UAV motion distortions and dynamic obstacles using optical flow with an improved segmentation network. \cite{Secchiero2024} combined semantic and stereoscopic information for SLZ identification post-exploration. However, significant challenges persist, especially in handling highly dynamic environments where high-risk obstacles like pedestrians and vehicles frequently and unpredictably alter the safety landscape. A common limitation in some systems is the simplification of risk assessment into binary classifications (safe/unsafe), which can obscure crucial nuances needed for decision-making when no perfectly risk-free area is available and a suitable alternative option must be selected. Furthermore, the ability to ``remember'' transient hazards—obstacles that move out of view but might reappear—is critical for conservative and truly safe landing decisions. This ``risk memory'' aspect, which prevents the system from being lured into areas that were recently dangerous, is often not explicitly addressed or robustly implemented. Moreover, a common trend in the literature is to focus primarily on the SLZ selection task, leaving the subsequent continuous guidance and control for the landing maneuver as a separate problem.
    
    Beyond general SLZ detection, research has also addressed specific categories of hazards or employed alternative strategies. For instance, avoiding human gatherings using crowd density maps \cite{Tzelepi2021,Gonzalez-Trejo2021_2} targets a particular risk factor. Others combine multiple sensor modalities, such as camera and LiDAR, to leverage complementary information for enhanced perception \cite{Lim2024}, or use object detectors in conjunction with occupancy maps to identify free space \cite{Mitroudas2024}. While interesting in their respective domains, these approaches might not offer the same level of comprehensive, scene-wide semantic risk assessment as a dedicated segmentation-based system or may rely on sensor fusion complexities. 
    
    Reinforcement learning (RL) has also been explored, often demonstrating success in navigating dynamic simulated environments for tasks like landing \cite{Vemulapalli2024}. However, many RL frameworks assume a pre-defined landing target or specific mission context (e.g., return-to-home) \cite{Liu2025}, which differs fundamentally from emergency scenarios requiring on-the-fly SLZ identification and assessment in unknown, potentially vast urban territory.
    
    The inherent difficulty and risk associated with evaluating such autonomous systems in live, populated urban settings under difficult real conditions has become one of the main challenges in developing and comparing fully autonomous landing solutions. Direct field tests in populated areas are often restricted by safety regulations, while purely virtual simulators (e.g., Gazebo, AirSim, Unreal) may struggle to replicate the visual richness and lighting conditions of real urban footage. Accordingly, some noteworthy evaluation frameworks have appeared very recently. These frameworks are essential for ensuring reproducibility and enabling rigorous, comparative testing. Prominent examples include platforms like VIVA-SAFELAND \cite{Soriano-Arxiv2025}, which our work leverages to conduct repeatable experiments using real urban aerial footage, and Robot-In-The-Loop (RITL) simulation pipelines \cite{Tovanche-Picon2024} that merge virtual environments with real hardware control.
    
    To our knowledge, a discernible gap exists in the literature for a fully autonomous, vision-only emergency landing system that can robustly and adaptively handle dynamic, complex urban environments without reliance on a priori known markers or pre-existing maps, by deeply integrating semantic risk assessment throughout the entire landing process. Our key contribution lies in a holistic, end-to-end solution that utilizes semantic segmentation to evaluate scene risk, select a suitable SLZ through a defined decision process, and execute the landing maneuver. This integrated methodology aims to significantly enhance UAV autonomy and safety during critical emergency landing situations in challenging real-world urban scenarios.

\section{Problem definition}
\label{sec:prob}

Consider a UAV equipped with a monocular camera $C$. Let the timeline be discretized in stages indexed as $t=0,1,\dots,k$, where $t=0$ refers to the time instant at which the emergency landing begins, and $k$ refers to the current time instant. We denote $I_t$ as the image taken with the camera $C$ at time $t$. Assume that a planar region $P \subset \mathbb{R}^2$ is a good enough approximation of the environment floor, which constitutes the set that contains the candidate SLZs for the UAV. The camera $C$ is pointing downward, so its optical axis is normal to $P$ using a stabilization device such as a gimbal or through software. Moreover, it is assumed that a computer vision module $\mathcal{M}_S$ is available, such that it can retrieve semantic information from an image $I_t$ relevant for landing. Lastly, it is considered that the center $\mathbf{c}$ of the image $I_t$ correctly models the center of the actual UAV landing region in the landing plane $P$. The objective is to design a fully autonomous landing solution that leverages semantic information from the currently available image set $\{I_t\}_{t=0}^k$, so that the solution includes SLZ identification until landing itself.  

\section{Proposed Risk-Aware Method}\label{sec:risk}
The risk-aware strategy presented in this article consists of several interconnected components that collectively enable safe autonomous landing in complex urban environments. This section provides a detailed examination of each component in our framework.
\subsection{Risk Map Generation}
\begin{figure}[!t]
    \centering
    \includegraphics[width=0.48\textwidth]{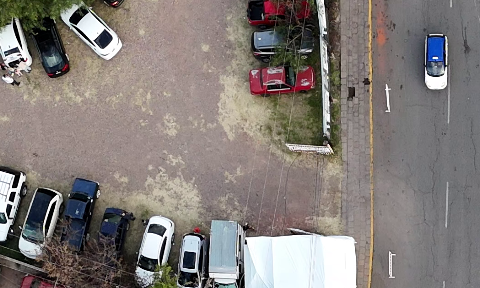}
    \vspace{0.2cm}
    \includegraphics[width=0.48\textwidth]{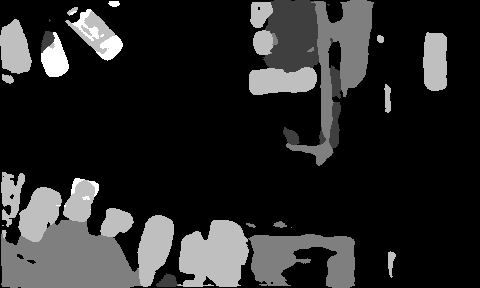}
    \caption{Left: Drone-view image $I_k$ from the UAV. Right: Grayscale representation of the risk map $\mathcal{R}_{k}$, where brighter pixels denote higher risk.}
    \label{fig:normalize}
\end{figure}

The proposed approach works online. The first step consists of applying a semantic segmentation to the current image $I_k$ using the computer vision module $\mathcal{M}_S$. 
The module $\mathcal{M}_S$ assigns each pixel to a category relevant to urban environments, such as grass, pavement, cars, people, vegetation, etc. Later, a risk mapping $\mathcal{M}_\mathcal{R}$ is applied such that each category is assigned a numerical risk value, with low values indicating safer terrain (e.g., grass $\rightarrow$ 0 or vegetation $\rightarrow$ 1) and high values indicating hazards (e.g., vehicles, pedestrians $\rightarrow$ 4) \cite{Loera2024}. Transforming segmentation labels into numeric scores produces a risk map $\mathcal{R}_{k} = \mathcal{M}_\mathcal{R}(\mathcal{M}_S(I_k))$. The objective of utilizing a risk map representation is to retrieve relevant safety information for emergency landing; this results in fewer classes that ease the decision-making process. 
Fig. \ref{fig:normalize} shows how a raw drone-view image $I_k$ is converted into a corresponding risk map $\mathcal{R}_{k}$.
\subsection{Risk memory}
Real urban scenes exhibit considerable variability due to pedestrians, vehicles, and lighting conditions. Semantic segmentation in such dynamic environments can often misclassify certain areas or yield inconsistent risk estimates between consecutive frames. To address this challenge, we implement a buffering mechanism that maintains a persistent ``memory'' of detected risks across frames. This memory is initialized at the beginning of the emergency landing procedure, that is, time $t=0$, and persists throughout the landing process.

\begin{figure}[!t]
    \centering
    \begin{tabular}{@{}c@{\hspace{0.1em}}c@{\hspace{0.1em}}c@{}}
        \includegraphics[width=0.32\textwidth]{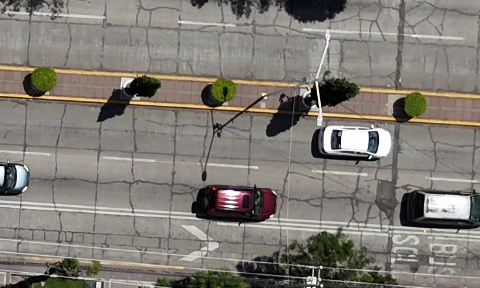} &
        \includegraphics[width=0.32\textwidth]{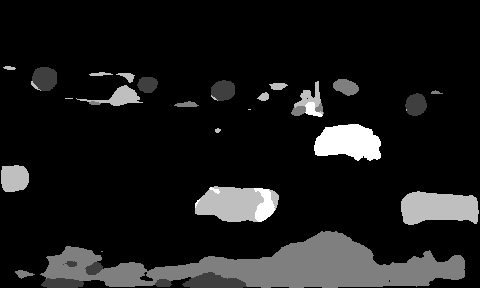} &
        \includegraphics[width=0.32\textwidth]{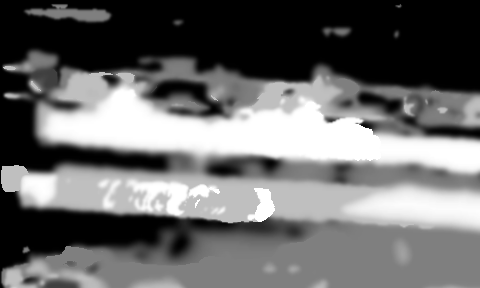} \\
        \includegraphics[width=0.32\textwidth]{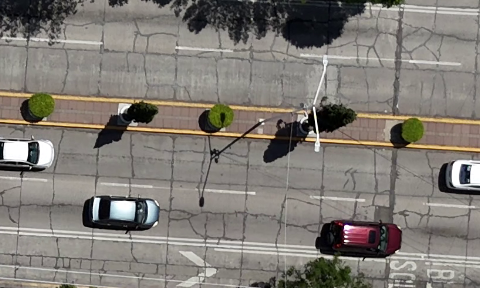} &
        \includegraphics[width=0.32\textwidth]{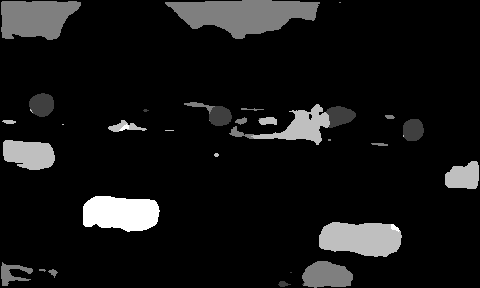} &
        \includegraphics[width=0.32\textwidth]{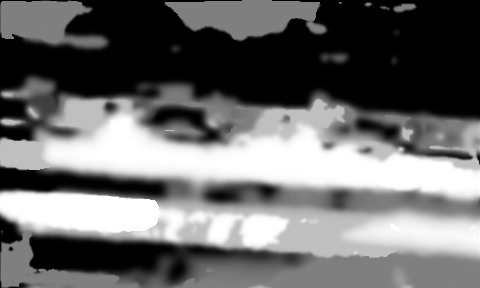} \\
        \footnotesize (a) & \footnotesize (b) & \footnotesize (c)
    \end{tabular}
    \caption{Two consecutive image frames demonstrating the \emph{risk memory} approach. Column (a) shows aerial drone views $I_{k-1}$ and $I_k$ with moving vehicles. Column (b) displays the raw risk maps $\mathcal{R}_{k-1}$ and $\mathcal{R}_{k}$ generated from semantic segmentation. Column (c) illustrates the corresponding local views 
    of the accumulated global risk maps $\mathcal{R}^{\text{global}}_{k-1}$ and $\mathcal{R}^{\text{global}}_{k}$ as seen from the drone's current perspective, showing how the pixel-wise maximum operation effectively preserves high-risk regions from previous frames within the drone's field of view.}
    \label{fig:memory}
\end{figure}

Rather than relying solely on frame-by-frame analysis, our approach maintains a global risk map that accumulates risk information up to time $k$. When a new frame $I_k$ is processed, we combine it with the existing global risk map $\mathcal{R}^{\text{global}}_{k}$ performing a pixel-wise maximum operation. This requires localizing the raw local risk map $\mathcal{R}_{k}$ within $\mathcal{R}^{\text{global}}_{k}$. To this end, an homography $\mathbf{H}_k$ can be obtained by means of the extrinsic parameters of the camera $C$, which are assumed to be retrievable by means of the UAV's on-board sensors. That is, the transformation between the coordinates that index the two risk maps is defined as 
\begin{equation} \label{eq:proj}
 \left[
   \begin{array}{c}
        x^{g}  \\
        y^{g} \\
        1   
\end{array}\right]
   =  \frac{1}{\lambda_k}\mathbf{H}_k
  \left[
  \begin{array}{c}
        x  \\
        y \\
        1   
\end{array}\right]
, 
\end{equation}
with $(x,y)$ and $(x^{g},y^{g})$ coordinates with respect to the reference frames that define $\mathcal{R}_{k}$ and $\mathcal{R}^{\text{global}}_{k}$, respectively, and $\lambda_k$ a scale factor that can be easily retrieved from the UAV altitude and the camera pose. Consequently, the global risk map $\mathcal{R}^{\text{global}}_{k}$ computation is as follows. 
\begin{equation} \label{eq:global_risk}
    \mathcal{R}^{\text{global}}_{k}(x^{g},y^{g}) = \max \left(\mathcal{R}^{\text{global}}_{k-1}(x^{g},y^{g}), \mathcal{R}_{k}(x,y) \right),
\end{equation}
for all $(x,y)$ that index $\mathcal{R}_{k}$, and with $(x^{g},y^{g})$ directly related to a point $(x,y)$ through (\ref{eq:proj}). Note that (\ref{eq:global_risk}) implicitly incorporates the risk information retrieved from the entire set of available images $\{I_t\}_{t=0}^k$ conservatively. 
%
This conservative approach offers two key advantages:
\begin{enumerate}
    \item \emph{Persistence of risk information:} Areas that were once identified as high-risk (such as roads where vehicles were detected) remain marked as risky even after the dynamic object moves away. This prevents the system from selecting landing zones in areas where vehicles and/or people circulate, which serves as a conservative prediction of risky future events.
    \item \emph{Robustness to segmentation variations:} Occasional false negatives in obstacle detection are mitigated, as a high-risk area correctly identified in any previous frame will be preserved in the global risk map. This helps to handle cases where obstacles are temporarily occluded or misclassified due to viewpoint or illumination changes.
\end{enumerate}
The resulting global risk map $\mathcal{R}^{\text{global}}_{k}$ provides a spatial record of hazards encountered throughout the flight, ensuring that transient dangers remain represented in the risk assessment even after they move or temporarily disappear from view. Fig.~\ref{fig:memory} illustrates this concept, showing how variations in consecutive risk maps are consolidated into a unified representation that preserves the highest risk detected at each location.
\subsection{Risk Expansion}
    We define $\mathcal{R}^{\text{local}}_{k} \subset \mathcal{R}^{\text{global}}_{k}$ as the portion of the global risk map $\mathcal{R}^{\text{global}}_{k}$ that falls within the drone's current field of view. Although this local view already incorporates persistent hazard information from the risk memory mechanism, landing zones selected directly from $\mathcal{R}^{\text{local}}_{k}$ may still be dangerously close to high-risk areas. To create safer margins, we implement an altitude-dependent risk expansion process that operates specifically on this local view, transforming it to better reflect safety requirements at different altitudes.

    At higher altitudes (above $30 m$), 
    we apply a large Gaussian filter directly to the local risk map $\mathcal{R}^{\text{local}}_{k}$. 
    At these elevations, fine-grained detail is less critical, as there is no immediate risk of collision, 
    allowing us to focus on finding suitable landing areas by smoothing local variations while preserving the general risk distribution patterns. The proposed risk expansion is applied as follows.
    %
    \begin{equation}
        \mathcal{R}^{\text{f}}_k = G_{\text{l}} * \mathcal{R}^{\text{local}}_{k},
    \end{equation}
where $\mathcal{R}^{\text{f}}_k$ denotes the filtered risk map after processing, $G_{\text{l}}$ indicates a Gaussian filter with a large kernel appropriate for high-altitude risk assessment, and $*$ represents the convolution operation. The size of this kernel is selected to balance computational efficiency with the need to exploit the UAV's high-altitude vantage point. 

    At lower altitudes (below $30 m$), 
    detecting small obstacles, such as pedestrians, is vital for safety. For effective risk expansion at these altitudes, we implement a process that begins with determining an appropriate dilation kernel size $k_d$ dependent on the drone's altitude; the value of $k_d$ decreases as the drone descends closer to the ground.
    %
    %
    The scaling of $k_d$ addresses two opposing factors that emerge as the UAV descends. On the one hand, the detection of critical obstacles like people becomes more reliable at lower altitudes due to their increased size in the field of view. On the other hand, semantic segmentation can actually become more imprecise when very close to the ground, with small noise artifacts being more prominent in the risk map. 
    After determining an adequate kernel size $k_d$ based on altitude, we apply a morphological dilation operation, $D_{k_d}(.)$, followed by a moderate-sized Gaussian filter $G_{\text{m}}$. The resulting filtered risk map is computed as 
    %
    \begin{equation}
        \mathcal{R}^{\text{f}}_k = G_{\text{m}} * D_{k_d}(\mathcal{R}^{\text{local}}_{k}).
    \end{equation}
    
\begin{figure}[!t]
        \centering
        \begin{tabular}{@{}c@{\hspace{0.1em}}c@{\hspace{0.1em}}c@{}}
            \includegraphics[width=0.32\textwidth]{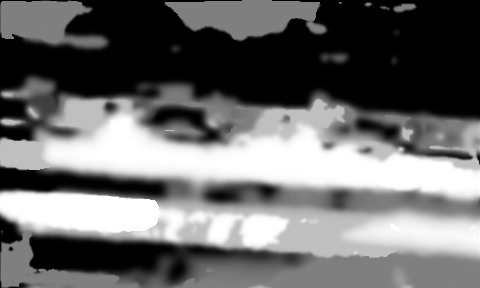} &
            \includegraphics[width=0.32\textwidth]{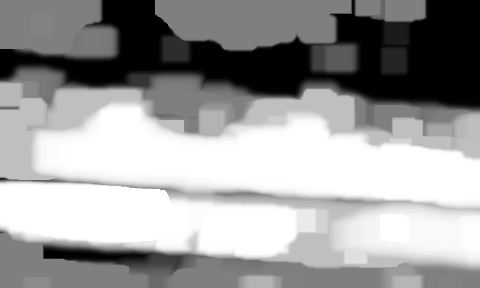} &
            \includegraphics[width=0.32\textwidth]{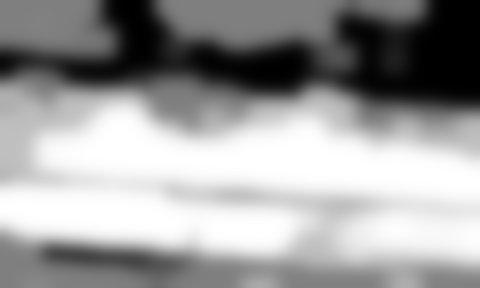} \\[0.6em]
            \includegraphics[width=0.32\textwidth]{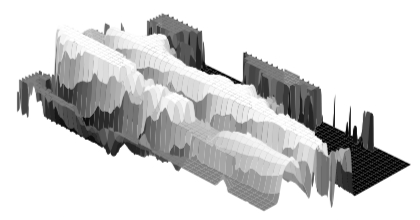} &
            \includegraphics[width=0.32\textwidth]{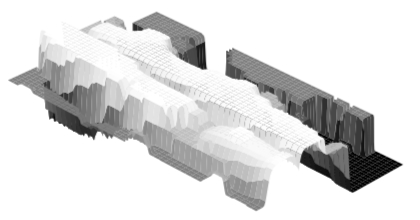} &
            \includegraphics[width=0.32\textwidth]{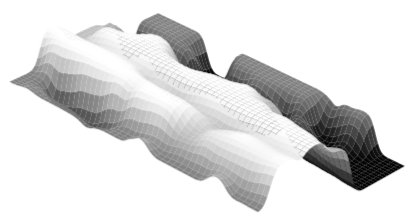}\\
            \footnotesize (a) & \footnotesize (b) & \footnotesize (c)
        \end{tabular}
        \caption{Progressive visualization of the risk expansion process. Top row shows 2D representations and bottom row shows corresponding 3D intensity profiles of the following. (a) Local view of the risk map $\mathcal{R}^{\text{local}}_{k}$. (b) Map after morphological dilation $D_{k_d}(\mathcal{R}^{\text{local}}_{k})$. (c) Final processed risk map after Gaussian filtering $\mathcal{R}^f_{k}$.}
        \label{fig:filtered}
    \end{figure}

    Fig.~\ref{fig:filtered} illustrates each stage of this process, showing the transformation from the local view of the global risk map through dilation to the final filtered risk map, which results in a smoother and more conservative map. 
\subsection{Local-Minimum Detection}
\begin{figure*}[!t]
        \centering
        \subfloat[]{\includegraphics[width=0.48\textwidth]{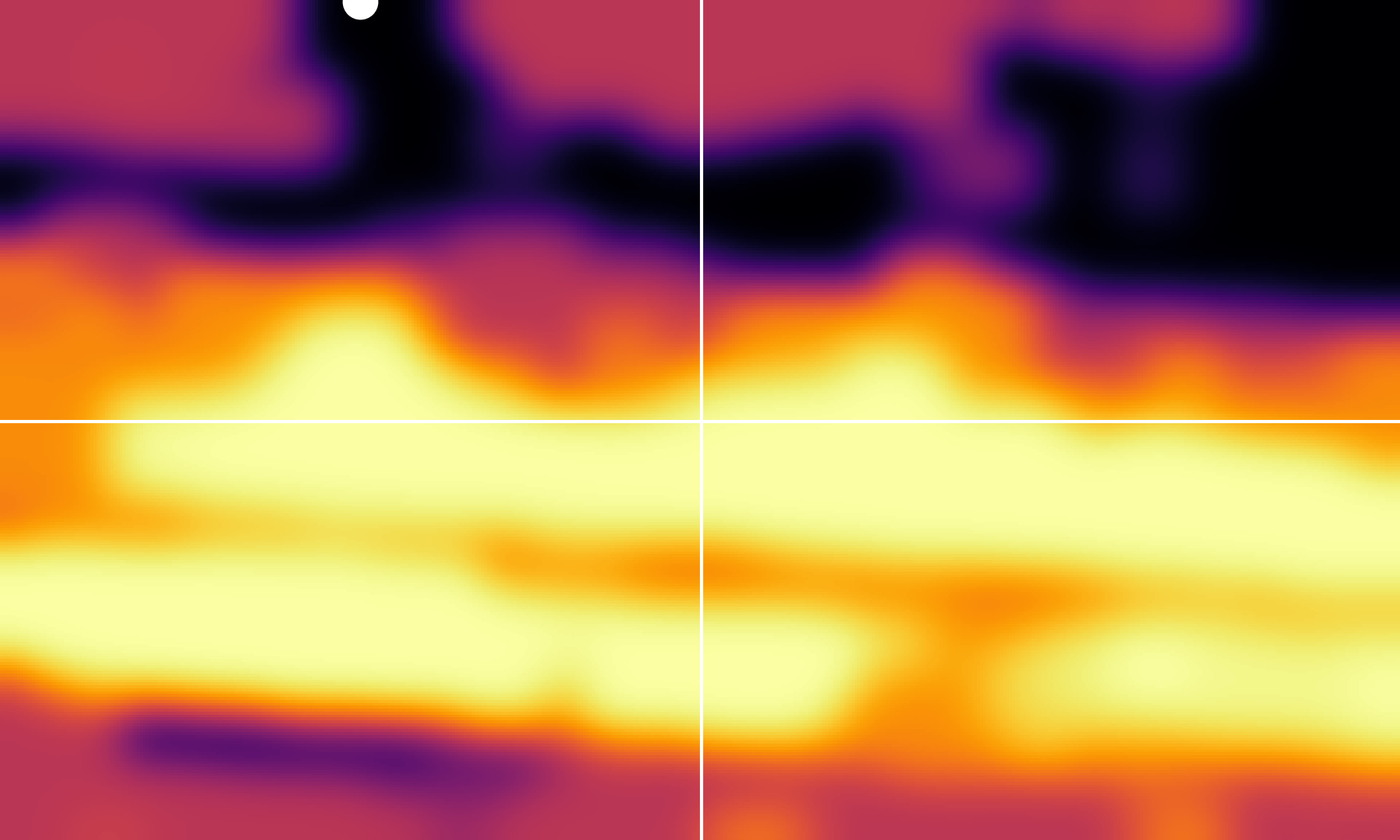}}
        \hfil
        \subfloat[]{\includegraphics[width=0.48\textwidth]{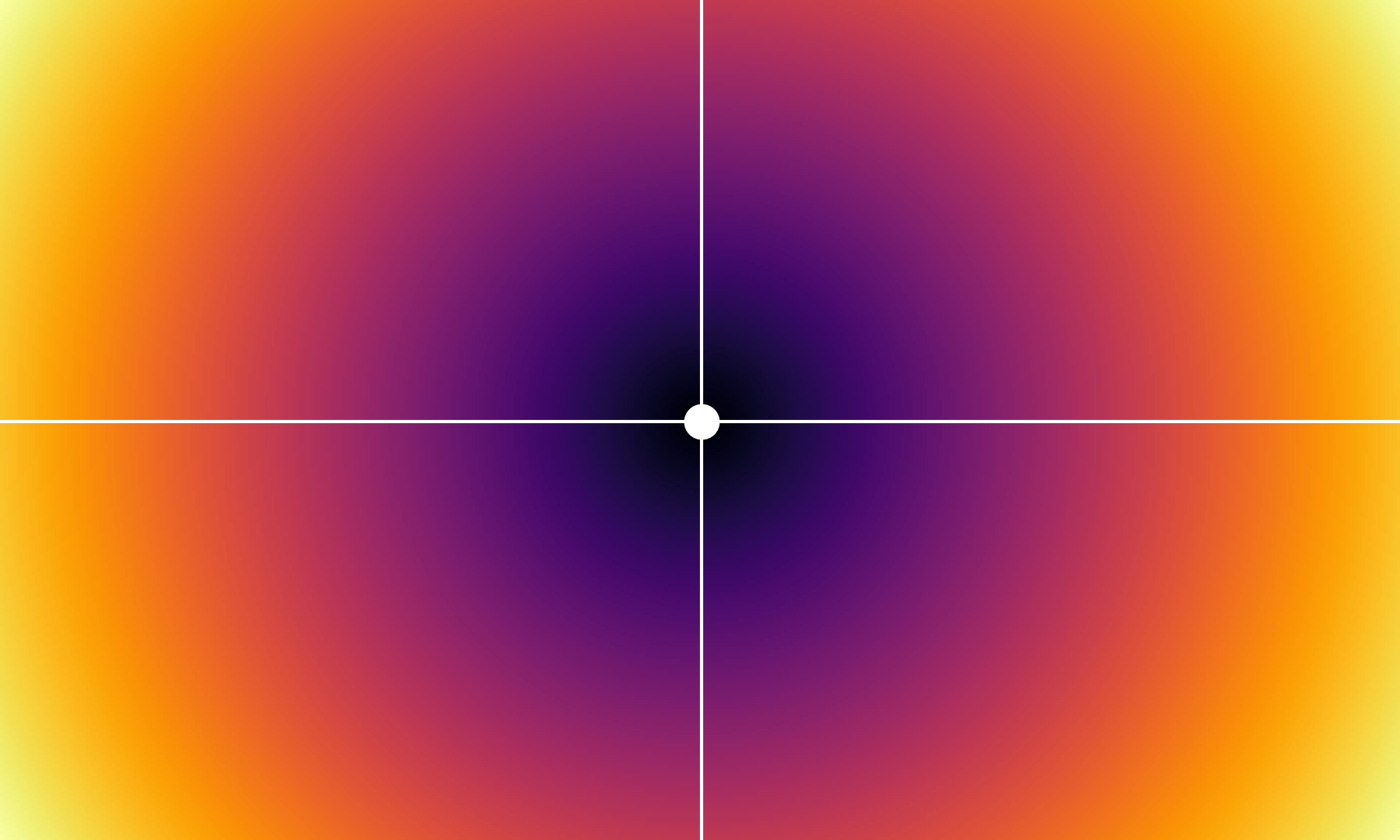}}
        \hfil
        \subfloat[]{\includegraphics[width=0.48\textwidth]{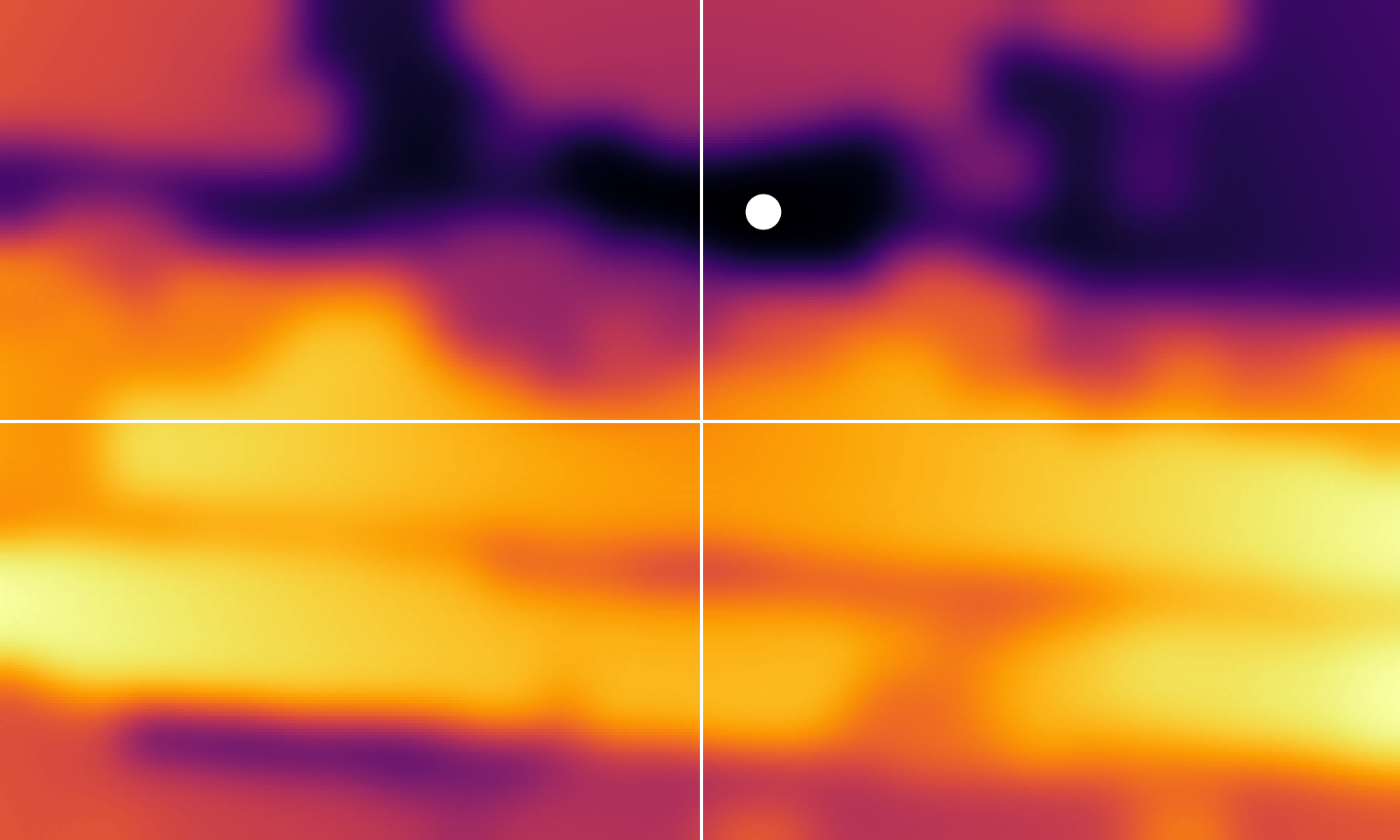}}
        \hfil
        \subfloat[]{\includegraphics[width=0.48\textwidth]{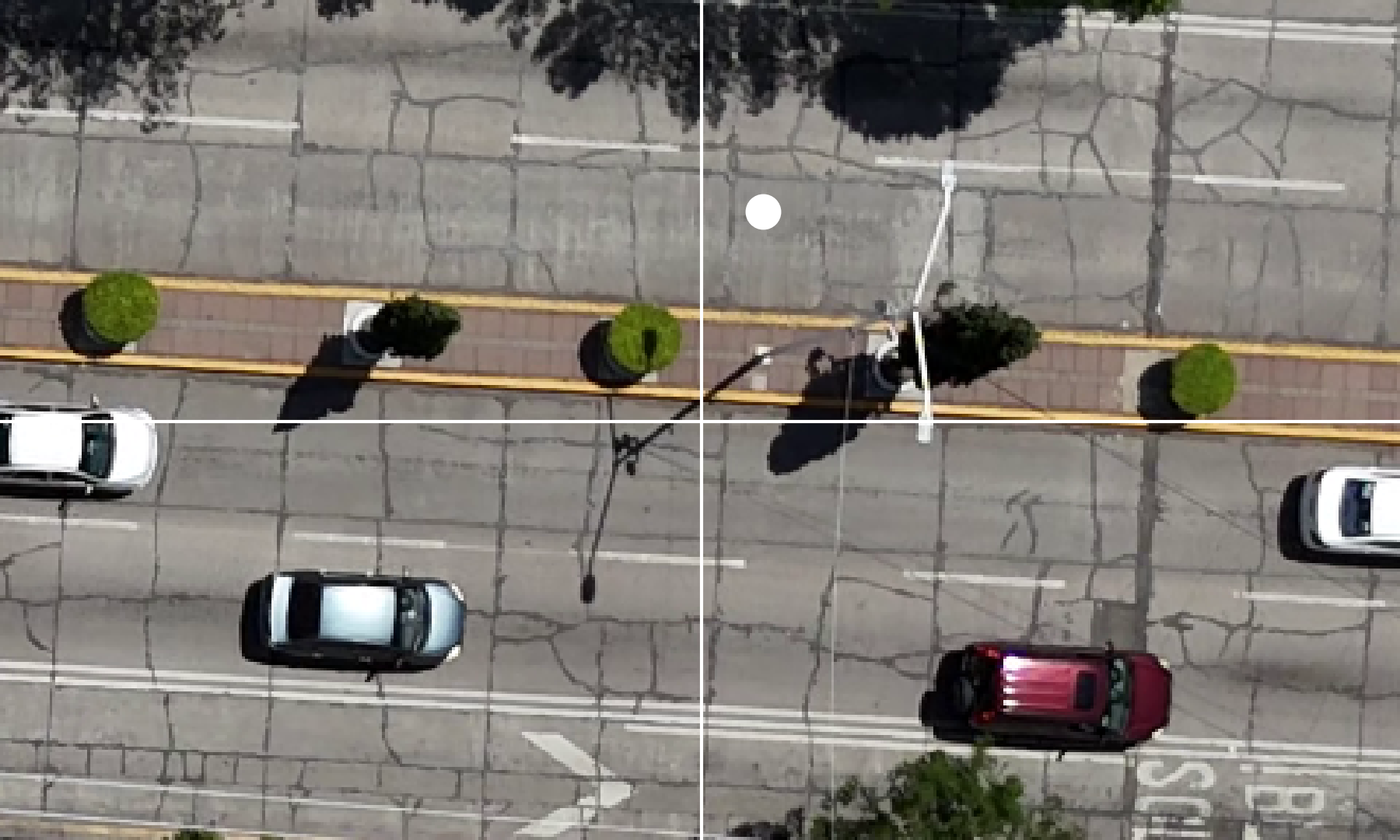}}
        \caption{Visualization of the landing point selection process. (a) Filtered risk map $\mathcal{R}^{\text{f}}_k$ with darker areas representing lower risk. (b) Distance map $\mathcal{L}$ centered at the image's optical center. (c) Weighted combination map $\mathcal{V}$ incorporating both risk and distance factors, with the white dot indicating the selected landing point $p^*_k$ that minimizes this combined cost. (d) Original drone view $I_k$ with the landing point projected onto the actual scene, where a road was selected, which, being free of traffic at that moment, constituted a valid and safe landing point.}
        \label{fig:localmin}
    \end{figure*}

    Once the filtered risk map $\mathcal{R}^{\text{f}}_k$ is obtained, we need to identify the optimal location for landing. The goal is to find a point that represents the ``lowest'' risk while also considering the practical constraints of the emergency landing scenario, for example, reaching the landing point quickly. 
    The proposed approach minimizes a weighted combination of the risk map and a proximity measure modeled as a distance map centered on the optical center of the image $I_k$, which is defined as
    \begin{equation}
        \mathcal{V}(p) = \alpha \cdot \mathcal{R}^{\text{f}}_k(p) + \beta \cdot \mathcal{L}(p, \mathbf{c}),
    \end{equation}
    where $\mathcal{L}(p,\mathbf{c})$ represents the Euclidean distance on the image plane from point $p$ to the image center $\mathbf{c}$, while $\alpha$ and $\beta$ are weighting coefficients that determine the relative importance of risk minimization versus distance minimization. The final landing point is then determined by finding the point $p_k^*$ that minimizes this local weighted map $\mathcal{V}$, that is,
    \begin{equation} \label{eq:pk}
        p^*_k = \arg\min_{p \in S(\mathcal{R}^{\text{f}}_k)} \mathcal{V}(p),
    \end{equation}
    with $S(\mathcal{R}^{\text{f}}_k)$ the set of coordinates that cover the filtered risk map $\mathcal{R}^{\text{f}}_k$. 
    The resulting low-risk point $p^*_k$ is taken as a candidate SLZ in the current frame $I_k$. Fig.~\ref{fig:localmin} illustrates how the approach highlights the final local minimum $p^*_k$.
\subsection{Landing point stabilization}
    For landing point stabilization, we employ a dual approach that addresses both spatial and temporal variability throughout the descent process. First, a queue of candidate landing points from the $N$ most recent image frames is maintained. Then, an average is performed on these candidates, as shown below.  
    \begin{equation}
        \overline{p^*_k} = \frac{1}{N}\sum_{i=0}^{N-1} p^*_{k-i},
    \end{equation}
    where $\overline{p^*_k}$ is the temporally averaged landing point at time $k$ and $p^*_{k-i}$ denotes the landing point at time $k-i$ selected through (\ref{eq:pk}). This temporal averaging reduces the impact of random noise and abrupt segmentation errors that occur in individual frames. 
    
    Once $\overline{p^*_k}$ is obtained, for a descent to go on, the center of the actual landing region of the UAV in the landing plane $P$ must remain for a given time period within a safety area $\mathcal{S}$, ensuring that the selected SLZ has temporal consistency. Considering that the image center $\mathbf{c}$ correctly models the center of the actual UAV landing region, the safety area $\mathcal{S}$ is projected onto the image plane as a circular region with radius $\tau(z)$ centered at $\mathbf{c}$. Therefore, the SLZ centered at $\overline{p^*_k}$ is said to be time consistent if in the image plane $\overline{p^*_k}$ remains within a distance $\tau(z)$ from $\mathbf{c}$ for at least $M$ time steps. Note that the radius $\tau(z)$ is a function of the UAV altitude $z$, since the computations are made in the image plane and the size of the projected safety area $\mathcal{S}$ increases as the UAV approaches the landing plane $P$. In practice, $\tau(z)$ is implemented using the camera model to convert a predefined radius in meters to pixels at the current altitude.

\begin{figure*}[!t]
        \centering
        \subfloat[]{\includegraphics[width=0.48\textwidth]{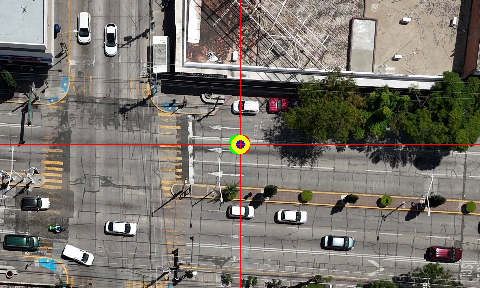}}
        \hfil
        \subfloat[]{\includegraphics[width=0.48\textwidth]{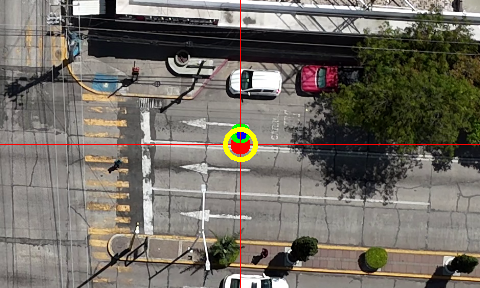}}
        \hfil
        \subfloat[]{\includegraphics[width=0.48\textwidth]{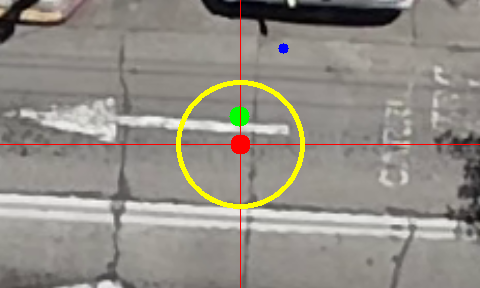}}
        \hfil
        \subfloat[]{\includegraphics[width=0.48\textwidth]{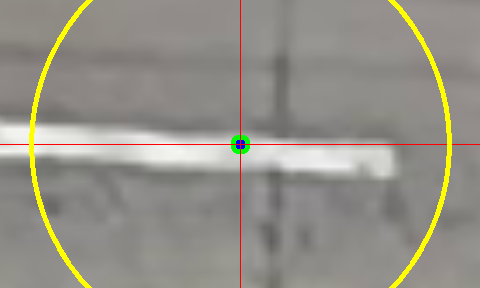}}
        \caption{Visualization of the landing point stabilization process at different altitudes. 
        The blue dot represents the instantaneous proposed landing spot $p^*_k$, the green dot shows the temporal averaging of landing points $\overline{p^*_k}$, the red dot marks the center $\mathbf{c}$ of image $I_k$, and the yellow circle is the the safety area $\mathcal{S}$. (a) Initial approach at higher altitude. (b) Descent phase showing the UAV's path toward the selected landing point. (c) and (d) lower altitude views.}
        \label{fig:stabilization}
    \end{figure*}

If at some time $k$ the UAV deviates from the safety zone $\mathcal{S}$, that is $\|\overline{p^*_k} - \mathbf{c}\| \geq \tau(z)$, the SLZ centered at $\overline{p^*_k}$ is no longer time consistent, the time accumulator is reset, and descent is paused until stability is regained. This approach ensures that landing proceeds only when consistently safe conditions are maintained. Fig.~\ref{fig:stabilization} demonstrates both mechanisms of our landing stabilization approach in action.

\subsection{Control and Navigation}
    After obtaining a robust estimate of the SLZ $\overline{p^*_k}$ through the stabilization mechanisms, we implement an image-based proportional-derivative (PD) control scheme to drive the UAV toward the target. This control approach translates the visual positional errors into appropriate flight commands.

    For horizontal positioning, reference roll and pitch angles, $\phi_c$ and  $\theta_c$, respectively, are controlled independently with PD controllers that are defined as
    \begin{align}
        \phi_c &= K_{p,x}\,\delta_x + K_{d,x}\,\dot{\delta}_x, \\ 
        \theta_c &= K_{p,y}\,\delta_y + K_{d,y}\,\dot{\delta}_y,
    \end{align}
    where $\delta_x$ and $\delta_y$ represent the horizontal errors of $\overline{p^*_k}$ from the image center $\mathbf{c}$, and $K_{p,\cdot}$ and $K_{d,\cdot}$ are proportional and derivative gains, respectively. These controllers generate roll and pitch commands that move the UAV horizontally to align with the selected landing zone. 
    
    For altitude control, we leverage the same altitude-dependent safety criterion established in the stabilization phase. Therefore, using the concept of the safety area $\mathcal{S}$ from the previous section, the desired altitude $z_d$ is decremented only when the UAV has maintained stable positioning above the selected SLZ for a sufficient period of time. That is, $z_d$ is computed as
    \begin{equation}
        z_d = 
        \begin{cases} 
          z_d - \Delta z & \text{if } \|\overline{p^*_k} - \mathbf{c}\| < \tau(z), \: \forall t > k - M,  \\
          z_d & \text{otherwise},
        \end{cases}
        \label{eq:th}
    \end{equation}
    with $\Delta z$ is the descent step. 
    Another PD controller then controls the vertical motion as
    \begin{equation}
        z_c = K_{p,z}(z_d - z) - K_{d,z}\dot{z},
    \end{equation}
    where $z_c$ is the altitude control command, $z$ is the current altitude, and $z_d$ is the desired altitude that decreases steadily during controlled descent.
    
    The control system guides the UAV to converge steadily above a consistently identified low-risk location before initiating and maintaining a controlled descent. The combination of vision-based risk assessment with responsive PD control allows the landing maneuver to adapt to dynamic urban environments with moving obstacles, while maintaining safety parameters throughout the entire process. Algorithm \ref{alg:safeLanding} summarizes our complete risk-aware landing approach, integrating all steps described in this section. 

\begin{algorithm}[!t]
    \caption{Risk-Aware Autonomous Landing}
    \label{alg:safeLanding}
    \begin{algorithmic}[1]
    \STATE \textbf{Input:} Drone-view image $I_k$, current altitude $z$
    \STATE \textbf{Output:} Control commands $\phi_c$, $\theta_c$, $z_c$
    \STATE $\mathcal{R}_{k} \gets \mathcal{M}_\mathcal{R}(\mathcal{M}_S(I_k))$; //\texttt{Apply semantic segmentation}
    \FORALL{$(x,y)$} 
    \STATE $\mathcal{R}^{\text{global}}_{k}(x^{g},y^{g}) \gets \max \left(\mathcal{R}^{\text{global}}_{k-1}(x^{g},y^{g}), \mathcal{R}_{k}(x,y) \right)$; //\texttt{Update global risk map using (\ref{eq:proj})}
    \ENDFOR
    \STATE $\mathcal{R}^{\text{local}}_{k} \gets \mathtt{RetrieveLocalView}(\mathcal{R}^{\text{global}}_{k})$; //\texttt{Get risk map within drone's field of view} 
    \IF{$z > 30m$}
        \STATE $\mathcal{R}^{\text{f}}_k \gets G_{\text{l}} * \mathcal{R}^{\text{local}}_{k}$; //\texttt{Apply large Gaussian filter}
    \ELSE
        \STATE $\mathcal{R}^{\text{f}}_k \gets G_{\text{m}} * D_{k_d}(\mathcal{R}^{\text{local}}_{k})$; //\texttt{Apply dilation and moderate-size Gaussian filter}
    \ENDIF
    \FORALL{$p \in S(\mathcal{R}^{\text{f}}_k)$}
    \STATE $\mathcal{V}(p) \gets \alpha \cdot \mathcal{R}^{\text{f}}_k(p) + \beta \cdot \mathcal{L}(p, \mathbf{c})$;  //\texttt{Compute weighted map}
    \ENDFOR
    \STATE $p^*_k \gets \arg\min_{p \in S(\mathcal{R}^{\text{f}}_k)} \mathcal{V}(p)$; //\texttt{Find minimum in weighted map and add to queue}
    \STATE $        \overline{p^*_k} \gets \frac{1}{N}\sum_{i=0}^{N-1} p^*_{k-i}$; //\texttt{Calculate temporal average of landing points}
    \IF{$\|\overline{p^*_k} - \mathbf{c}\| < \tau(z), \: \forall t > k - M$}
    \STATE $z_d \gets z_d - \Delta z$; //\texttt{Update desired altitude}
    \ENDIF
    \STATE $(\delta_x, \delta_y) \gets \overline{p^*_k} - \mathbf{c}$; //\texttt{Horizontal errors}
    \STATE //\texttt{Compute control commands}
    \STATE $\phi_c \gets K_{p,x}\,\delta_x + K_{d,x}\,\dot{\delta}_x$; 
    \STATE $\theta_c \gets K_{p,y}\,\delta_y + K_{d,y}\,\dot{\delta}_y$;
    \STATE $z_c \gets K_{p,z}(z_d - z) - K_{d,z}\dot{z}$;
    \RETURN $\phi_c$, $\theta_c$, $z_c$
    \end{algorithmic}
\end{algorithm}

\section{Experimental Validation}\label{sec:experiment}
    To evaluate the effectiveness of our risk-aware landing approach, we conducted testing using diverse real urban aerial environments through the ViVa-SAFELAND framework \cite{Soriano-Arxiv2025}, which leverages real aerial videos to simulate landing missions for an emulated aerial vehicle (EAV). 
    This allows us to randomize several trials and capture different quantitative metrics. Also, this tool enables the reproducibility of test scenarios for a fair comparison between different vision-based landing solutions. The test urban scenarios include environments with dynamic obstacles, such as moving vehicles and pedestrians, and different ground textures captured at multiple altitudes. This structured approach allows us to assess the method's performance using real urban footage without the safety risks of physical drone testing in populated areas, thereby providing meaningful insights into the algorithm's potential for real-world applications.
    %
    All tests were performed using an AMD Ryzen 7 5700X processor, 32 GB of RAM, and a NVIDIA GeForce RTX 3060 graphics card. The landing proposal was implemented using Python. In particular, the segmentation vision module $\mathcal{M}_S$ was implemented using U-Net.
    %
%
\subsection{Metrics for Landing System Evaluation}
    To comprehensively evaluate our vision-based landing system, we employ four complementary metrics that assess different aspects of the landing process. Since no absolute ground truth exists for landing zones in real urban footage, we establish a reference standard using a virtual ``metric camera'' $C_M$ positioned at a fixed height of $30m$ above the scene. Importantly, the camera $C_M$ remains horizontally aligned with the EAV's current position at all times, maintaining the same $x-y$ coordinates but at a constant altitude, regardless of the EAV's actual height. 
    In addition, the $30m$ altitude represents a nice balance where semantic segmentation has been observed to function reliably, capturing sufficient details while maintaining a broad perspective. As the EAV descends to lower altitudes where segmentation quality might degrade, the camera $C_M$ continues to provide a stable reference view. 
    
    For our evaluation purposes, areas of low risk, such as grass or pavement, are considered valid landing zones that do not pose safety concerns. Using $C_M$ as our reference, we compute the following evaluation metrics.
    
    \emph{- Success Metric.} This metric determines whether a landing spot is free of collision with risky obstacles by evaluating a circular safety area of $0.5m$ radius around the target point. It calculates whether any pixels classified as ``high-risk'' exist within this circle; that is, if any pixel is over a risk threshold $\gamma_r$, as viewed from the metric camera. A landing is considered successful only if no high-risk pixels are present in the landing area. This is quantified as
    %
    \begin{equation}
        M_{suc} = \begin{cases} 
            0 & \text{if } \exists p \in \mathcal{C}_{0.5} : \mathcal{R}_{C_M}(p) > \gamma_r,\\
            1 & \text{otherwise},
        \end{cases}
    \end{equation}    
    where $\mathcal{C}_{0.5}$ represents the set of pixels within the circular area, and $\mathcal{R}_{C_M}(p)$ is the risk value of pixel $p$ as captured by the metric camera $C_M$. This strict binary criterion ensures that only completely safe areas without any high-risk elements are considered successful landing zones.\\
    \emph{- Risk Metric.} 
    Using a larger $2m$ diameter circular area, this metric calculates the proportion of high-risk pixels within this extended zone, which is computed as
    \begin{equation}
        M_{risk} = \frac{\sum_{p \in \mathcal{C}_2} \boldsymbol{1}(\mathcal{R}_{C_M}(p) > \gamma_r)}{|\mathcal{C}_2|},
    \end{equation}
    where $\mathcal{C}_2$ represents the set of pixels within the 2-meter diameter circle as viewed from $C_M$, $|\mathcal{C}_2|$ denotes the total number of pixels in this area, and $\boldsymbol{1}(\cdot)$ returns 1 if its argument is true, and 0 otherwise. A lower Risk Metric value indicates a safer landing area with fewer high-risk elements in the vicinity.
    
    \emph{- Proximity Metric (Nearest Obstacle).}
    The Proximity Metric offers a continuous measure of safety by quantifying the distance between the landing point and the closest potential hazard. Using the $C_M$ camera view, this metric calculates the minimum distance from the center of the landing area to any high-risk pixel, and is calculated as
    \begin{equation}
        M_{prox} = \min_{p \in \mathcal{H}} \|p - g\|_2,
    \end{equation}
    where $\mathcal{H}$ is the set of high-risk pixels (with values $> \gamma_r$) in the view of the metric camera $C_M$, $g$ is the center of the landing area, again in the $C_M$ view, and $\|\cdot\|_2$ represents the Euclidean distance. Higher values indicate more favorable landing zones with greater clearance from obstacles, providing an important safety margin measurement.
    
    \emph{- Warning Metrics.} To further characterize the safety margins around potential landing zones, we introduce two complementary warning metrics that build upon the Proximity Metric by categorizing obstacles according to their distance from the landing spot. Their computation is as follows. 
    \begin{equation}
        W_i = 
            \begin{cases}
                1 & \text{if } \min_{p \in \mathcal{H}} \|p - g\|_2 < \omega_i, \\
                0 & \text{otherwise},
            \end{cases} 
    \end{equation}
   where each $W_i \in\{W_1,W_2\}$ has its corresponding warning zone according to a safety radius $\omega_i$ that triggers the  warning. Warning $W_1$ identifies critical proximity violations where high-risk elements are detected within $1 m$ of the landing point, representing an immediate hazard. Warning $W_2$ identifies secondary proximity violations where high-risk elements are between $1 m$ and $2 m$ from the landing spot, representing a potential concern but with a modest safety margin. As expected, it is not desirable for any warnings to be triggered during landing.    
    
    \emph{- IoU Metric (Intersection over Union).} The IoU metric evaluates the semantic segmentation network consistency between the risk perception at the EAV's variable altitude and the reference view at $30m$. This spatial overlap measurement provides critical insight into segmentation reliability during descent operations. For each frame during the landing process, we calculate the pixel-wise agreement between high-risk areas identified from the current viewpoint and those from the metric camera's fixed perspective using the standard intersection over union formula
    \begin{equation}
        \text{IoU} = \frac{|\mathcal{P} \cap \mathcal{G}|}{|\mathcal{P} \cup \mathcal{G}|},
    \end{equation}
    where $\mathcal{P}$ represents the set of pixels predicted as high-risk from the EAV's current view $\mathcal{R}_{k}$, and $\mathcal{G}$ represents the set of pixels labeled as high-risk in the $C_M$ view $\mathcal{R}_{C_M}$.
    
    This metric is tracked continuously once the EAV descends below $30 m$, and the final reported value represents the average IoU across this entire descent phase. This metric, therefore, provides valuable insights into how reliably the system maintains accurate risk perception at different altitudes where landing decisions become increasingly critical.\\
    
    \emph{- Execution Time Metric.} This metric quantifies the temporal efficiency of the landing process by measuring the total duration from when the landing zone search begins until the EAV completes its descent, that is,
    \begin{equation}
        t_{\text{total}} = t_{\text{end}} - t_{\text{init}},
    \end{equation}
    where $t_{\text{init}}$ marks the moment when the emergency landing task starts, and $t_{\text{end}}$ represents the moment when the EAV completes its landing. It is important to highlight that the complete landing strategy is performed in real-time within the VIVA-SAFELAND framework. By comparing execution times, 
    this metric helps evaluate the practical feasibility of the approach, particularly in time-sensitive scenarios where both safety and speed are important considerations, for instance, in the case of battery shortage.
\subsection{Test Environments and Scenarios}
To thoroughly evaluate our approach, we selected four distinct urban environments that present various challenges for safe drone landing (see Table \ref{tab:metrics_results}). 
%
%
For each environment, we tested our algorithm, running $100$ trials, randomizing the starting EAV position. Two landing approaches are tested, a baseline where the EAV randomly selects its landing position according to a uniform distribution, which is further referred to as \emph{uncontrolled}, and our risk-aware landing proposal, which is referred to as \emph{controlled}. The following four different configurations are tested for each environment: 
\begin{itemize}
    \item \textbf{SU}: Static-Uncontrolled. Using the same static picture, but without active landing zone selection, serves as a baseline comparison using default random positioning.
    \item \textbf{SC}: Static-Controlled. Using a single static picture, the landing algorithm actively selects a suitable SLZ in a static scene. 
    \item \textbf{DU}: Dynamic-Uncontrolled. Using the complete video sequence without active landing zone detection, we provided a reference baseline with random landing positions to evaluate the benefits of our active control approach.
    \item \textbf{DC}: Dynamic-Controlled. Using the complete dynamic video, the landing algorithm continuously identifies suitable SLZ and attempts to land under real dynamic conditions.
\end{itemize}
These configurations allow us to quantify the improvements provided by our active detection and landing system compared to naive positioning approaches, both in static images and dynamic video sequences. Table~\ref{tab:metrics_results} shows some statistics for the 100 trials corresponding to percentage ratios or mean values. As demonstrated in Table~\ref{tab:metrics_results}, the controlled trials (SC and DC) significantly outperformed their uncontrolled counterparts (SU and DU) across all safety-related metrics, with success rates reaching $90\%$-$100\%$ in controlled trials compared to only $45\%$-$86\%$ in the uncontrolled ones. Uncontrolled approaches also exhibit higher risk percentages and closer proximities to obstacles, particularly in the dynamic uncontrolled (DU) configuration, which represents the most challenging scenario. Our controlled approach achieves superior safety results while still operating in real-time, with complete landing maneuvers consistently executed in a reasonable time, demonstrating its practical viability for emergency landing scenarios in complex urban environments. A video showcasing the performance of the landing algorithm in different challenging urban scenarios is provided at \url{https://youtu.be/uZEgroQpMSI}.
%
%
\begin{table}[!t]
\footnotesize
    \caption{Performance Metrics for Different Urban Environments ($\uparrow$ the bigger the better, $\downarrow$ the smaller the better)}
    \label{tab:metrics_results} \renewcommand{\arraystretch}{1.2} 
    \centering
    \begin{tabular}{cccccccccc}
        \toprule
        \textbf{Scene} & \textbf{} & \textbf{Succ}  & \textbf{Risk }  & \textbf{Prox} & \textbf{W1} & \textbf{W2}  & \textbf{IoU}  & \textbf{Time}  \\
        
        \textbf{} & \textbf{} & \textbf{[\%]} $\uparrow$ & \textbf{[\%]} $\downarrow$ & \textbf{[m]} $\uparrow$ & \textbf{[\%]} $\downarrow$ & \textbf{[\%]} $\downarrow$ & $\uparrow$ & \textbf{[s]} $\downarrow$ \\
        \bottomrule
        \multirow{4}{*}{\includegraphics[height=2.0cm]{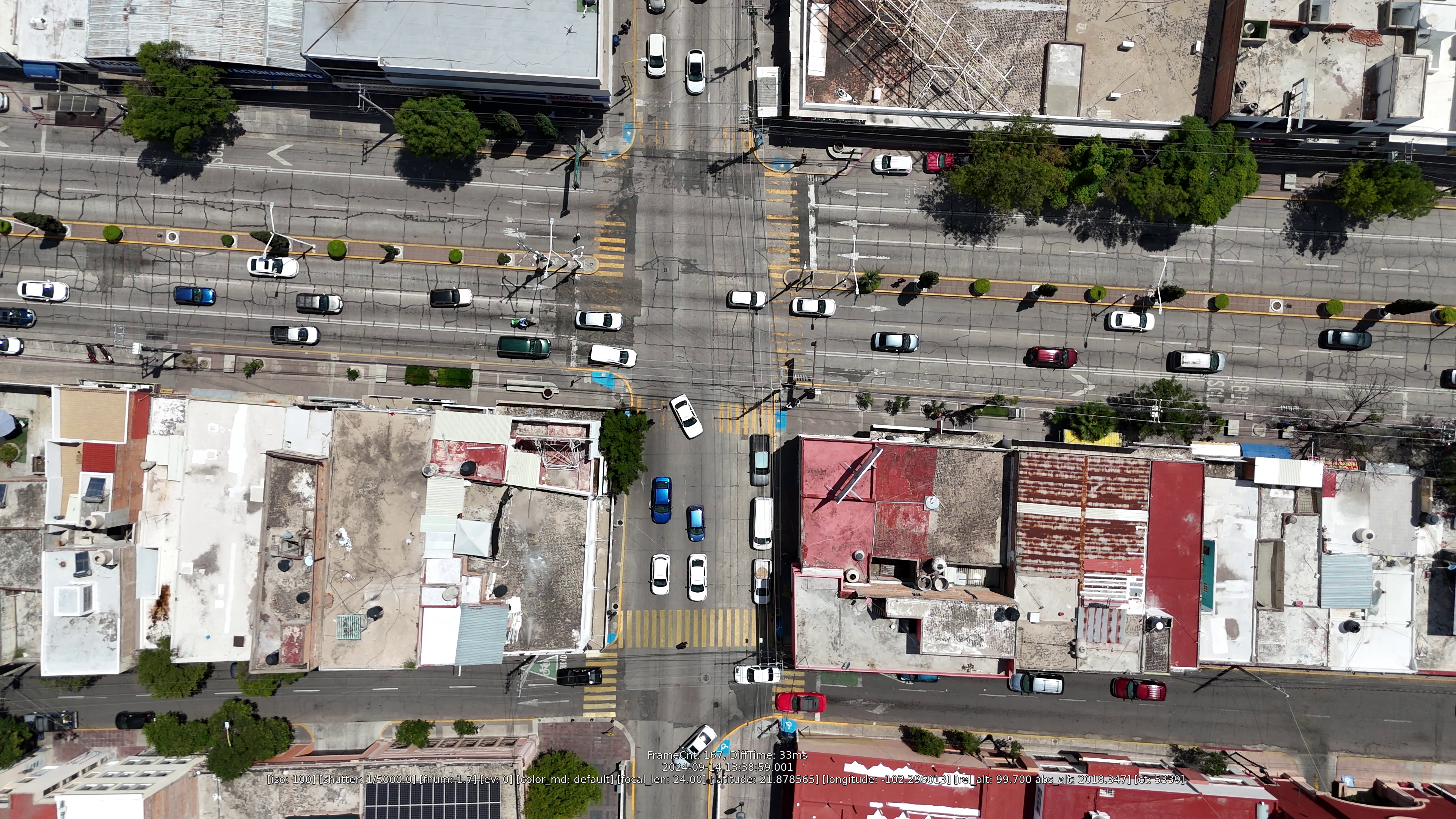}}
            & SU & 69.0 & 19.00 & 2.05 & 37.0 & 20.0 & 0.31 & \textbf{5.20} \\
            & SC & \textbf{98.0} & \textbf{0.856} & \textbf{3.50} & \textbf{3.00} & \textbf{17.0} & \textbf{0.42} & 12.57 \\
            & DU & 54.0 & 32.21 & 1.66 & 52.0 & \textbf{15.0} & \textbf{0.47} & \textbf{6.63} \\
            & DC & \textbf{93.0} & \textbf{3.22} & \textbf{2.49} & \textbf{8.00} & 25.0 & 0.43 & 17.22 \\
        \bottomrule
        \multirow{4}{*}{\includegraphics[height=2.0cm]{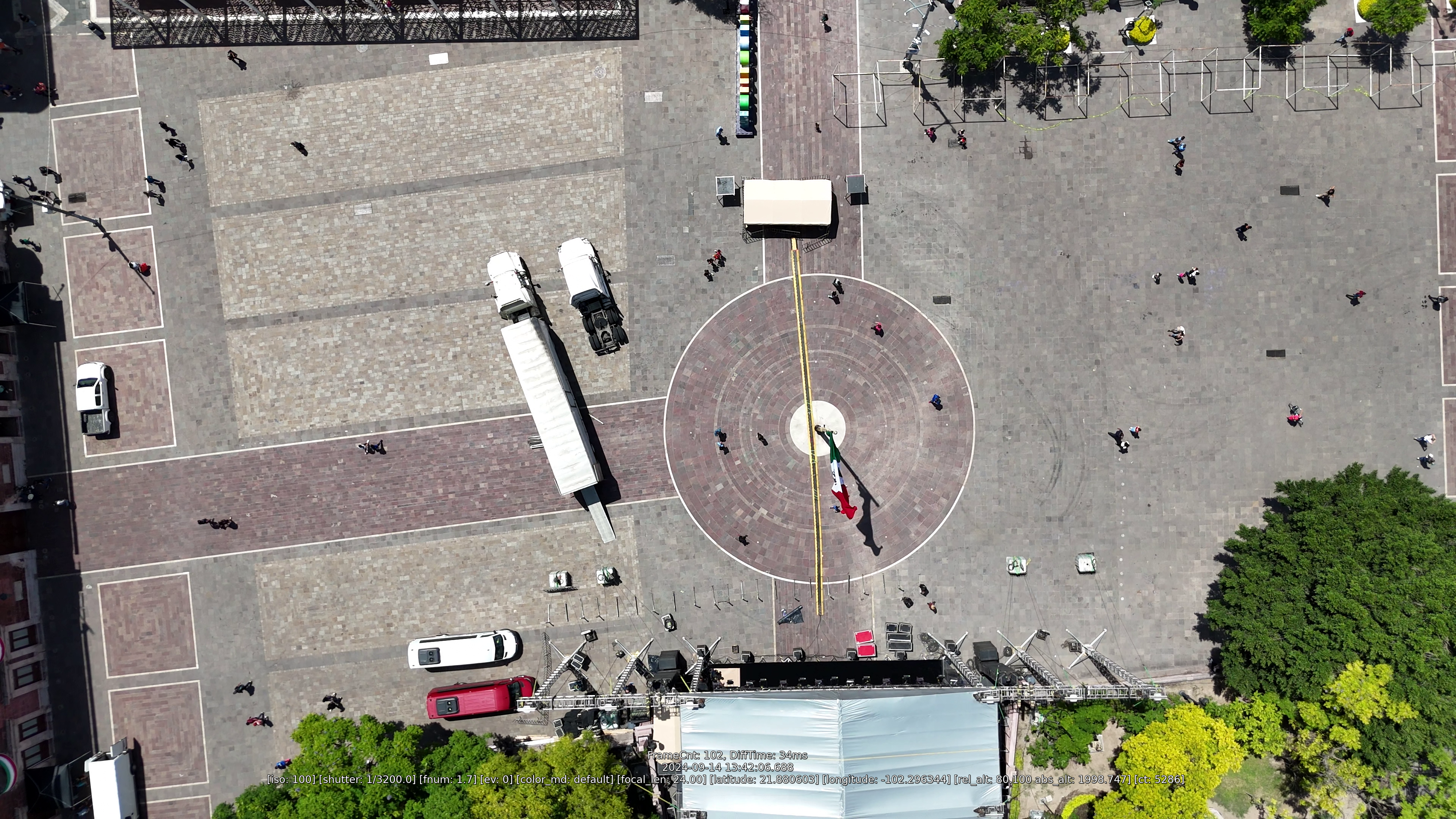}}
            & SU & 86.0 & 7.39 & 3.88 & 19.0 & 14.0 & 0.26 & \textbf{4.96} \\
            & SC & \textbf{100} & \textbf{0.14} & \textbf{4.76} & \textbf{2.00} & \textbf{3.00} & \textbf{0.43} & 12.14 \\
            & DU & 63.0 & 24.64 & 2.48 & 43.0 & \textbf{11.0} & 0.40 & \textbf{6.88} \\
            & DC & \textbf{100} & \textbf{0.22} & \textbf{3.99} & \textbf{4.00} & 6.00 & \textbf{0.41} & 15.20 \\
        \bottomrule
        \multirow{4}{*}{\includegraphics[height=2.0cm]{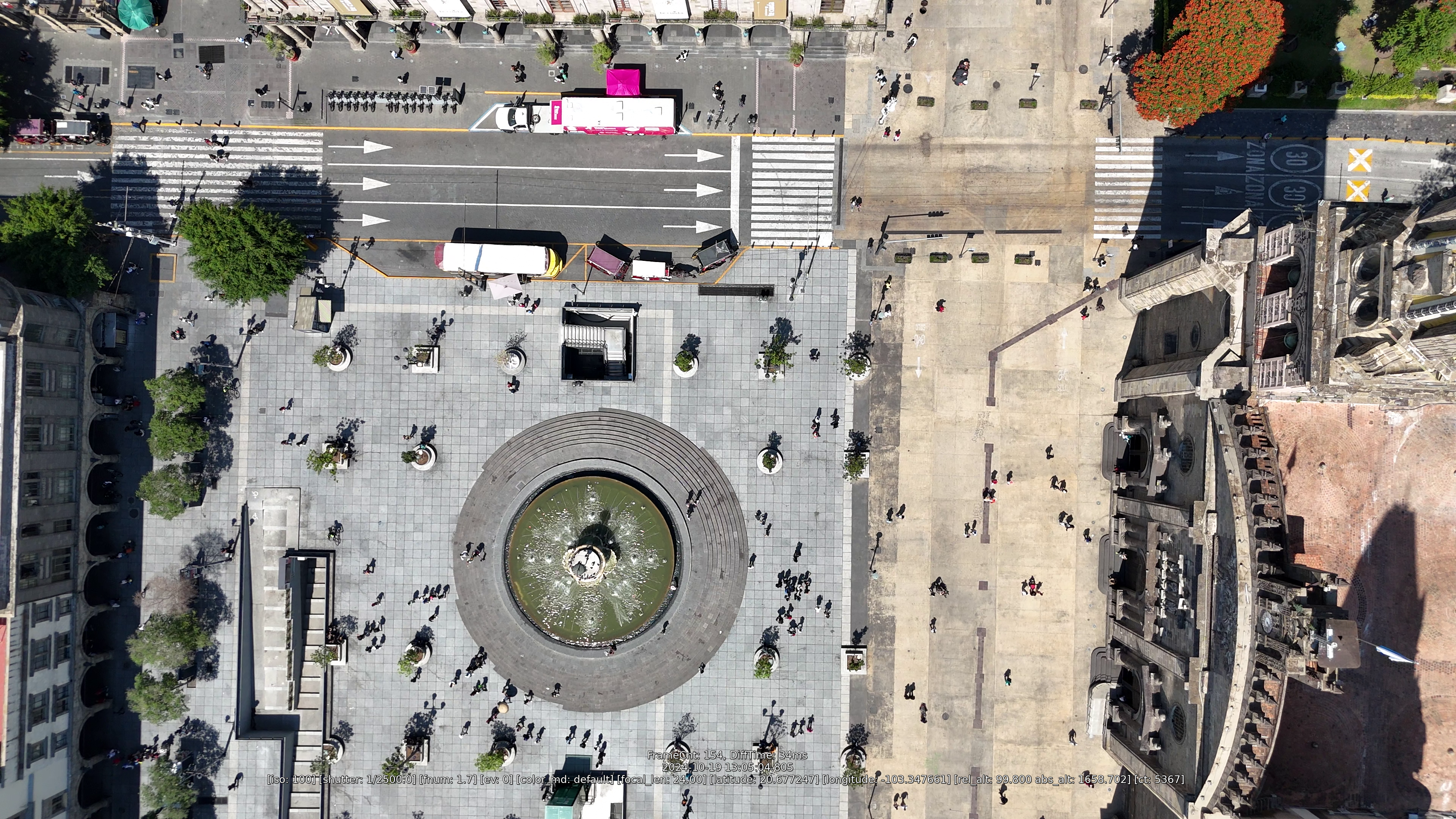}}
            & SU & 80.0 & 11.08 & 2.96 & 23.0 & 18.0 & 0.33 & \textbf{5.24} \\
            & SC & \textbf{100} & \textbf{0.11} & \textbf{4.29} & \textbf{2.00} & \textbf{6.00} & \textbf{0.37} & 11.66 \\
            & DU & 66.0 & 25.43 & 1.58 & 49.0 & \textbf{16.0} & \textbf{0.43} & \textbf{6.80} \\
            & DC & \textbf{96.0} & \textbf{1.02} & \textbf{3.84} & \textbf{5.00} & 9.00 & 0.40 & 15.80 \\
        \bottomrule
        \multirow{4}{*}{\includegraphics[height=2.0cm]{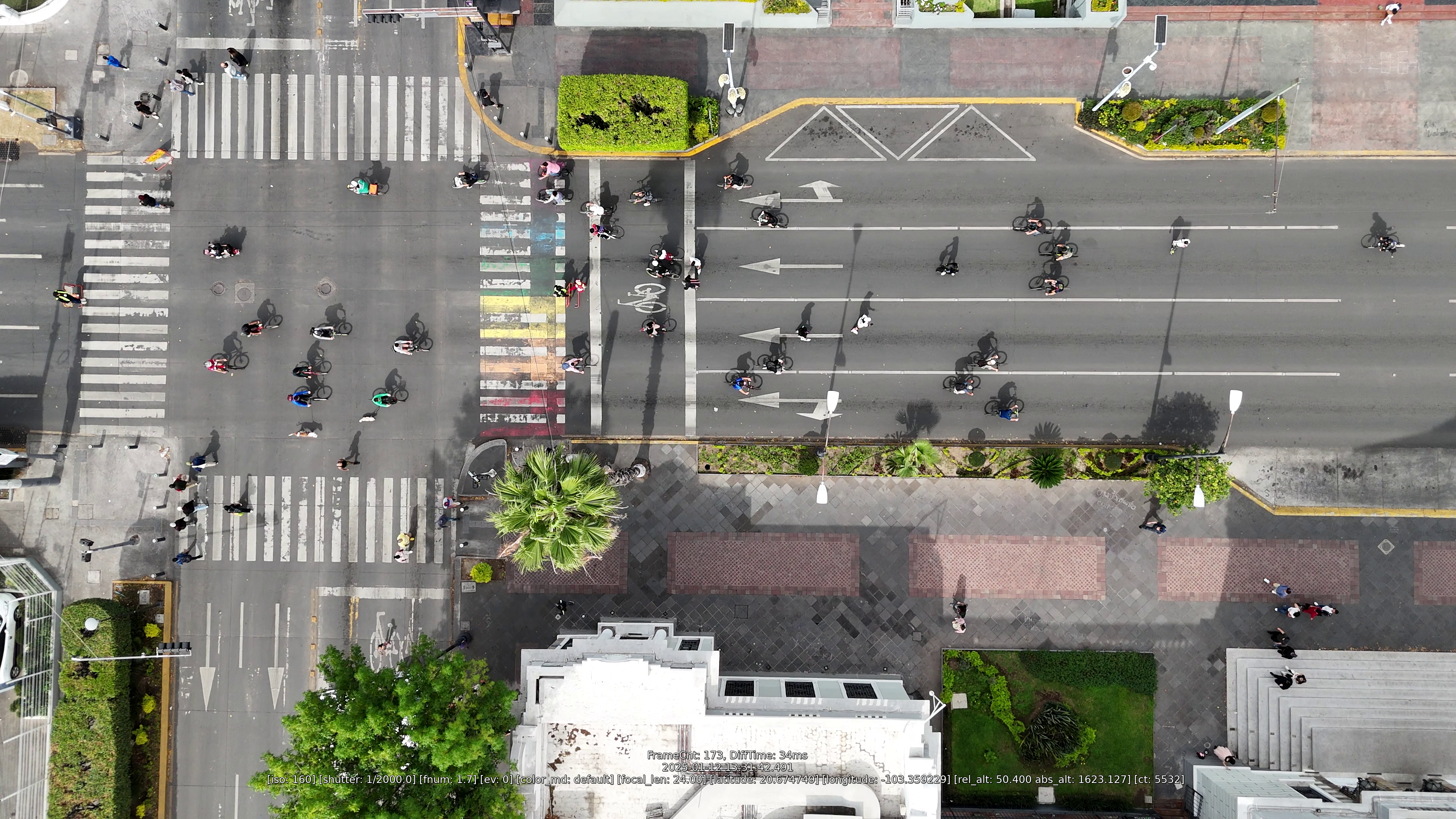}}
            & SU & 70.0 & 17.54 & 2.33 & 37.0 & 21.0 & 0.26 & \textbf{4.930} \\
            & SC & \textbf{100} & \textbf{0.02} & \textbf{3.59} & \textbf{0.00} & \textbf{4.00} & \textbf{0.39} & 10.95 \\
            & DU & 45.0 & 41.03 & 0.94 & 69.0 & \textbf{11.0} & 0.42 & \textbf{6.35} \\
            & DC & \textbf{90.0} & \textbf{5.030} & \textbf{2.10} & \textbf{20.0} & 26.0 & \textbf{0.44} & 15.24 \\
        \bottomrule
    \end{tabular}
\end{table}
\subsection{Metric Camera Validation}
While the metric camera $C_M$ at $30 m$ altitude provides a useful reference for quick evaluation, it relies on the same semantic segmentation network as the EAV, potentially propagating similar biases or errors. To address this limitation and more rigorously validate our approach, we conducted an additional evaluation using manually annotated ground truth data where high-risk elements were carefully labeled frame by frame in a test video sequence. Table \ref{tab:gt_comparison} compares our metric camera-based evaluation and the manually labeled ground truth, running $100$ randomized trials under the same video. As expected, the IoU metric, which evaluates the quality of the segmentation network, deteriorates when using the manually labeled validation video; nevertheless, we can observe that the main risk metrics are comparable in both validations, suggesting that the use of the $C_M$ camera provides an acceptable means for rapid assessment in various real scenarios, without the need for labor-expensive manual labeling of the videos frame by frame.
\begin{table}[!t]
    \caption{Comparison Between Metric Camera and Ground Truth Evaluation ($\uparrow$ the bigger the better, $\downarrow$ the smaller the better)}
    \label{tab:gt_comparison}
    \renewcommand{\arraystretch}{1.2}
    \centering
    \begin{tabular}{@{}lcc@{}}
        \toprule
        \textbf{Metric} & \textbf{ Camera $C_M$} & \textbf{Ground Truth} \\
        \midrule
        Success Rate [\%] $\uparrow$& \textbf{98.0} & \textbf{98.0} \\
        Risk [\%] $\downarrow$ & 1.38 & \textbf{1.19} \\
        Proximity [m] $\uparrow$& 4.22 & \textbf{6.07} \\
        Warning 1 [\%] $\downarrow$ & 8.0 & \textbf{2.0} \\
        Warning 2 [\%] $\downarrow$ & 10.0 & \textbf{3.0} \\
        IoU $\uparrow$ & \textbf{0.67} & 0.53 \\
        Execution Time [s] $\downarrow$ & 28.89 & \textbf{23.72} \\
        \bottomrule
    \end{tabular}
\end{table}
The results show that our metric-camera assessment consistently overestimates risk relative to the manually labeled ground truth, a conservative bias that, far from being detrimental, means the evaluation errs on the side of caution and that the true safety margins are probably even larger than those reflected in our main evaluation metrics.
\section{Conclusion and Future Work}\label{sec:conclusion}

This paper presented a vision-based risk-aware control approach for safe landing in complex urban environments. By leveraging semantic segmentation to create pixel-wise risk maps and applying conservative altitude-dependent risk propagation techniques, the system reliably avoids hazardous zones at various flight phases. The global risk map approach maintains a persistent ``memory'' of detected hazards, while temporal averaging of candidate landing points stabilizes detection and reduces transient errors. Our local-minimum detection component nicely balances risk minimization with practical distance considerations, identifying suitable landing zones even in complex environments with varying risk distributions. A PD control system continuously guides the UAV towards the identified SLZ, while our safety-first descent strategy initiates the actual landing process only after sufficient criteria have been maintained.

 Our experimental results on real scenes demonstrate the effectiveness of the proposed method, with controlled strategies consistently achieving success rates of $90\%$-$100\%$ compared to only $45\%$-$86\%$ in uncontrolled trials. Notably, in the most challenging urban environment with high pedestrian density and unpredictable movement patterns (Scene 4), our controlled approach achieved an outstanding $90\%$ success rate compared to just $45\%$ in the uncontrolled case. 
The proposed strategy proved to be effective in reducing the risk of accidents during real-time emergency landing, achieving successful landings in less than $20s$. To our knowledge, this is the first time that a quantitative evaluation is performed in a purely visual-based landing approach in complex urban scenarios, hence providing a fair baseline for comparison with future works.


For future work, we plan to implement this approach on real UAV platforms under controlled conditions, gradually transitioning from simplified environments to more complex urban settings as safety protocols permit. Additionally, we aim to integrate online object tracking for enhanced dynamic obstacle avoidance, estimate obstacles, depth, explore end-to-end deep learning architectures for direct SLZ inference, and refine vertical descent strategies once the UAV is centered over the selected landing zone. These advancements will further strengthen the practical applicability of risk-aware landing systems in real-world emergency scenarios.
 \bibliographystyle{elsarticle-num} 
  \bibliography{output}
\end{document}